\documentclass{bmvc2k}

\title{Camera-Agnostic Pruning of 3D Gaussian Splats via Descriptor-Based Beta Evidence}

\addauthor{Peter Fasogbon}{peter.fasogbon@nokia.com}{1}
\addauthor{Ugurcan Budak}{ugurcan.budak.ext@nokia.com}{1}
\addauthor{Patrice Rondao Alface}{patrice.rondao_alface@nokia.com 
}{2}
\addauthor{Hamed R. Tavakoli}{hamed.rezazadegan_tavakoli@nokia.com}{1}

\addinstitution{
	Nokia Technologies,\\
	Espoo, Finland
}
\addinstitution{
	Nokia Technologies, \\
	Antwerp, Belgium
}

\runninghead{Fasogbon et al.}{Camera-Agnostic Pruning of 3D Gaussian Splats}

\usepackage{graphicx}  % for \resizebox
\usepackage{multirow}  % for \multirow

\usepackage{amsmath,amssymb}

\let\originallesssim\lesssim
\let\originalgtrsim\gtrsim

\DeclareRobustCommand{\lesssim}{%
	\mathrel{\mathpalette\lowersim\originallesssim}%
}
\DeclareRobustCommand{\gtrsim}{%
	\mathrel{\mathpalette\lowersim\originalgtrsim}%
}

\makeatletter
\newcommand{\lowersim}[2]{%
	\sbox\z@{$#1<$}%
	\raisebox{-\dimexpr\height-\ht\z@}{$\m@th#1#2$}%
}
\makeatother

\begin{document}
	
	\maketitle
	
	\begin{abstract}
		The pruning of 3D Gaussian splats is essential for reducing their complexity to enable efficient storage, transmission, and downstream processing. However, most of the existing pruning strategies depend on camera parameters, rendered images, or view-dependent measures. This dependency becomes a hindrance in emerging camera-agnostic exchange settings, where splats are shared directly as point-based representations (e.g., \texttt{.ply}). In this paper, we propose a \emph{camera-agnostic}, \emph{one-shot}, \emph{post-training} pruning method for 3D Gaussian splats that relies solely on attribute-derived neighbourhood descriptors. As our primary contribution, we introduce a hybrid descriptor framework that captures structural and appearance consistency directly from the splat representation. Building on these descriptors, we formulate pruning as a statistical evidence estimation problem and introduce a \emph{Beta evidence} model that quantifies per-splat reliability through a probabilistic confidence score.
		
		Experiments conducted on standardized test sequences defined by the ISO/IEC MPEG Common Test Conditions (CTC) demonstrate that our approach achieves substantial pruning while preserving reconstruction quality, establishing a practical and generalizable alternative to existing camera-dependent pruning strategies.
	\end{abstract}
	
	%-------------------------------------------------------------------------
	\section{Introduction}
	\label{sec:intro}
	
	3D Gaussian Splatting (3DGS) has recently emerged as a powerful representation for real-time novel view synthesis, achieving high visual fidelity by modelling scenes as collections of anisotropic Gaussian primitives~\cite{kerbl2023gaussians}. By enabling fast rasterization and
	high-quality rendering, 3DGS has quickly become a competitive alternative
	to neural radiance fields (NeRFs) \cite{mildenhall2020nerfrepresentingscenesneural} for interactive applications. However,
	realistic scenes often require hundreds of thousands to millions of
	splats, leading to substantial storage costs, increased memory
	footprint, and higher computational overhead. These characteristics
	limit the practical deployment of 3DGS models on resource-constrained
	platforms such as mobile devices, embedded systems, and bandwidth-limited
	streaming scenarios.

	\vspace{0.1cm}
	
	To address this challenge, pruning and compression of Gaussian splats
	have become active research topics. A prominent post-training
	prune-recover methodology in LightGaussian~\cite{fan2024lightgaussiann}
	estimates global splat significance, performs spherical harmonics
	distillation, and applies vector quantization as part of an integrated
	compression pipeline. This approach achieves substantial model size
	reduction while largely preserving rendering quality. Similarly,
	Confident Splatting~\cite{razlighi2025confidentsplatting} introduces
	confidence-aware pruning by learning per-splat Beta distributions during
	training, using view-dependent reconstruction signals to guide opacity
	modulation and pruning decisions. MaskGaussian~\cite{Liu_2025_CVPR} further advances pruning with probabilistic masks and masked rasterization, enabling currently unused Gaussians to keep receiving gradients and thereby improving adaptive importance estimation during optimization. Despite their effectiveness, these approaches fundamentally rely on training-time supervision and camera-dependent information. Such assumptions do not hold in emerging MPEG standardization activities, notably within the Immersive 3D Gaussian Splatting (I-3DGS)
	paradigm~\cite{mpegi3dgs}, where Gaussian splats are exchanged directly as
	geometry-based point representations (e.g., \texttt{.ply} files) without
	access to training images, camera intrinsics, or extrinsics. In these
	interchange scenarios, pruning must operate purely post-training and
	remain fully camera-agnostic.

	%Both LightGaussian and Confident Splatting rely on rendered images, per-view visibility, or gradients accumulated during training, implicitly assuming access to camera intrinsics, extrinsics, and view trajectories. 
	
	\vspace{0.cm}
	
	\begin{figure*}
		\begin{center}
			\fbox{ \includegraphics[height=6.5cm, width=12.0cm]{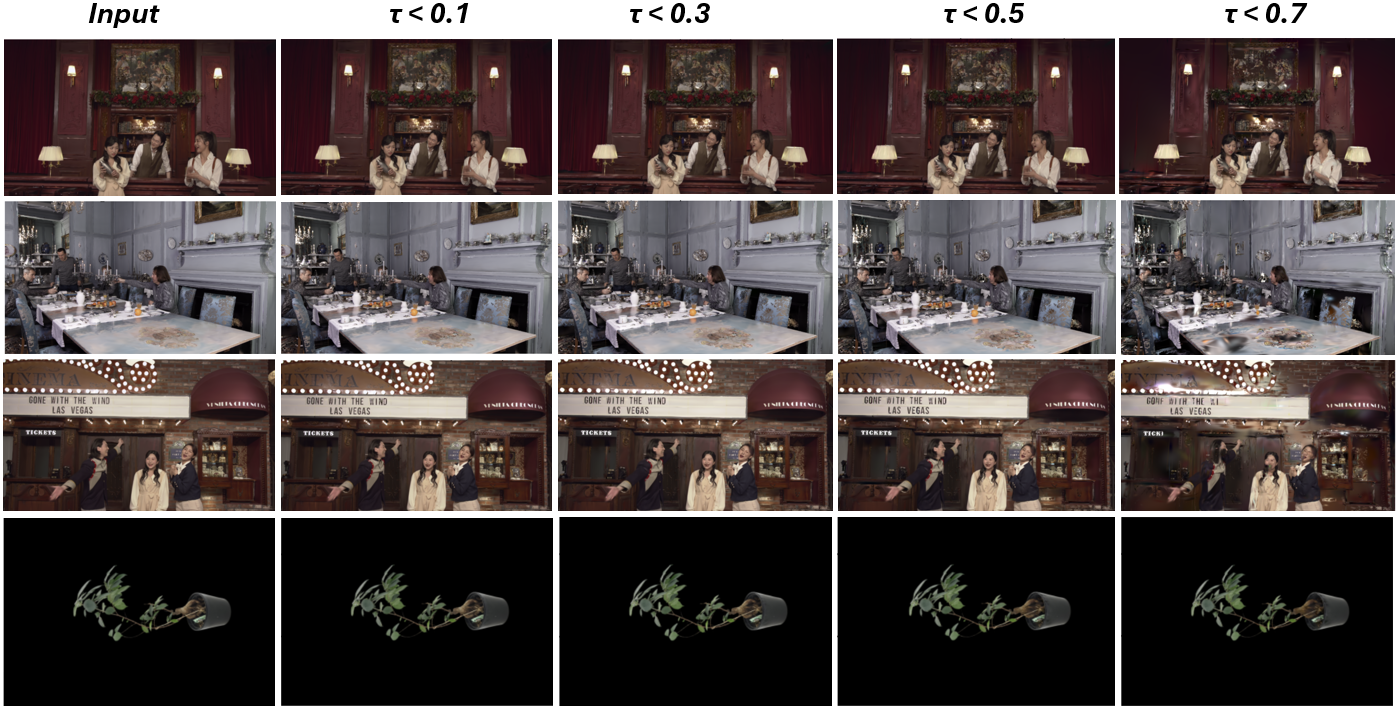}}
		\end{center}
		\caption{
			Overview of the proposed Gaussian splat pruning method on an example scene from the ISO/IEC MPEG CTC dataset~\cite{mpegdatasets}, rendered from a fixed camera viewpoint. The image columns show the non-pruned baseline and increasing pruning thresholds $\tau=0.1, 0.3, 0.50,$ and $0.7$, which remove about $10\%$, $30\%$, $50\%$, and $70\%$ of splats, respectively. 
		}
		\label{fig:main}
	\end{figure*}

	Unlike Confident Splatting, where Beta distribution parameters are learned jointly with splat parameters during training, our approach estimates Beta evidence entirely \emph{post-training} from descriptor-derived statistics. We formulate pruning as a statistical evidence estimation problem and derive per-splat confidence from Hybrid Splat Feature Histogram (HSFH) descriptors computed directly from the learned Gaussian representation. This design ensures full compatibility with MPEG I-3DGS interchange pipelines while preserving important structures under aggressive redundancy removal. Figure~\ref{fig:main} illustrates the qualitative effect on a representative CTC scene~\cite{mpegdatasets}: as the pruning threshold increases, progressively more low-confidence splats are removed while visual quality remains largely preserved across moderate thresholds $\tau$, with noticeable degradation only under aggressive pruning.

	\section{Background and Related Work}
	
	%\subsection{Properties of 3D Gaussian Splatting Representations}
	
	3DGS~\cite{kerbl2023gaussians} models a scene as a set of anisotropic Gaussian primitives parameterized by 3D position, covariance, opacity, and view-dependent appearance (e.g. spherical harmonics). The rendering is performed via differentiable rasterization and alpha compositing, enabling real-time performance while maintaining high visual fidelity. Unlike classical point clouds, each Gaussian splat carries anisotropic spatial support and learned radiance parameters. Therefore, splat importance is not purely geometric and thus depends jointly on its covariance structure, opacity, and appearance. This coupling between geometry and appearance complicates redundancy estimation compared to traditional mesh or point-based simplification. Furthermore, 3DGS splats are optimized for rendering rather than explicit geometric reconstruction. As a result, heuristics designed for point decimation or mesh simplification do not directly transfer to Gaussian-based representations.

	\subsection{Pruning in 3DGS}
	
	Existing pruning strategies for 3DGS can be categorised by the technique used to estimate splat importance.
	\vspace{-0.3cm}
	\paragraph{Training-Stage Pruning.}
	The original 3DGS framework~\cite{kerbl2023gaussians} prunes Gaussians during optimization based on opacity thresholds, excessive scale growth, or negligible gradient contribution. Since these criteria depend on rendering loss and view visibility, they are inherently tied to training trajectories. Subsequent methods, such as BOGausS~\cite{Pateux2025BOGausS}, track training-stage behaviours to improve optimization efficiency while implicitly removing unimportant splats. Mini-Splatting~\cite{Fang2024MiniSplatting} and EAGLES~\cite{Girish2024EAGLES} prioritize Gaussians using visibility and contribution statistics accumulated from training views. RAP~\cite{yang2026rapfastfeedforwardrenderingfree} trains a lightweight MLP during optimization to infer per-Gaussian confidence scores, while using rendering supervision to guide importance estimation. These approaches remain tightly coupled to gradient-based optimization and camera-dependent rendering signals.

	\paragraph{Gradient-Based and Learned Pruning.}
	More recent methods introduce differentiable or learned importance estimation. LP-3DGS~\cite{LP3DGS2024} learns a binary pruning mask via a Gumbel-Sigmoid relaxation, requiring iterative joint optimization. PUP-3DGS~\cite{PUP3DGS2025} employs second-order sensitivity analysis based on Hessian approximations to score Gaussians. MaskGaussian~\cite{Liu_2025_CVPR} models Gaussian existence with probabilistic masks and proposes masked rasterization so that even currently unused Gaussians can still receive gradients, mitigating irreversible pruning decisions caused by deterministic masks. Opacity-assisted pruning~\cite{HighFold2025} and semantic-guided approaches such as Clean-GS~\cite{CleanGS2026} similarly depend on training-time gradients, visibility statistics, or segmentation masks. Although these strategies achieve strong compression-performance trade-offs, they require access to training views, camera parameters, or optimization signals.

	\paragraph{Post-Training Compression Pipelines.}
	Post-training compression has also been explored in the literature. LightGaussian~\cite{fan2024lightgaussiann} combines global significance-based pruning with spherical harmonics distillation and vector quantization. While effective, its significance estimation and recovery stages rely on rendering evaluation driven by camera parameters. Confident Splatting~\cite{razlighi2025confidentsplatting} introduces a probabilistic confidence model by jointly learning per-splat Beta distributions during training. This provides principled uncertainty-aware pruning, but requires modifying the training procedure and leveraging view-dependent reconstruction signals. Recent works such as MesonGS~\cite{fu2024mesongspostinferencecompression} and ReduceGS~\cite{bruel2025reducegs} also operate post-training, but target attribute transform / vector quantization and hierarchical multi-scale simplification respectively. These methods are complementary to, not competitors of the proposed method, and could in principle be applied as a \emph{post-processing} to our method in a full compression pipeline. %We therefore keep our experimental comparison focused on methods that share the same interchange setting (pruning-only, unquantized \texttt{.ply}).

	\paragraph{Structural and Feature-Space Pruning.}
	Some approaches target latent or structural components instead of splats directly. Hash Grid Feature Pruning~\cite{HashGridPruning2025} removes unused hash-grid feature entries by analyzing geometric support from Gaussian centers. Adversarial Pruning Networks (APNet)~\cite{adversarialprunning} apply adversarial learning to prune Gaussians in a data-driven manner. These methods reduce redundancy in auxiliary representations but either rely on supervised retraining or do not operate directly on the Gaussian primitives themselves.
	
	\paragraph{NeRF and Visibility-Based Pruning.}
	Related work in NeRF compression prunes network weights, hash grids, or spatial regions based on ray-based saliency. Examples include coreset-based neuron pruning~\cite{CoresetNeRF2025}, HollowNeRF~\cite{HollowNeRF2023}, and visibility-driven geometry pruning~\cite{VisibilityPruning2024}. While conceptually related, these methods operate on neural field representations and depend on ray sampling or view statistics.

	\subsection{Intrinsic Cues from Local Structure}
	
	When importance must be inferred from the Gaussian representation itself, redundancy estimation must rely on intrinsic structural cues. In classical 3D vision, local reliability is commonly derived from neighborhood descriptors computed over point clouds or surface patches. Hand-crafted descriptors such as Spin Images~\cite{johnson1999spinimages}, SHOT~\cite{tombari2010shot}, and Fast Point Feature Histograms (FPFH)~\cite{rusu2009fpfh} encode geometric consistency within local neighborhoods and have been widely used for registration and robustness analysis. In addition, learning-based point descriptors, including PointNet-family encoders~\cite{qi2017pointnetdeephierarchicalfeature} and contrastive local embeddings~\cite{xiong2020locolocalcontrastiverepresentation}, further demonstrate that neighbourhood structure provides informative signals even in the absence of camera information.
	
	However, Gaussian splats differ fundamentally from raw points: they incorporate anisotropic covariance, opacity, and radiance parameters optimized for differentiable rendering. Descriptor-driven reliability estimation tailored specifically to Gaussian primitives remains largely unexplored. This motivates this work to propose splat-aware neighbourhood descriptors that jointly capture geometric and appearance consistency.

	\subsection{Positioning of This Work}
	
	Our work addresses the post-training regime in which only a finalized set of Gaussian splats (e.g., a \texttt{.ply} asset) is available, without access to cameras, rendered images, or training-time gradients. In contrast to camera-dependent prune-and-recover pipelines~\cite{fan2024lightgaussiann}, training-coupled confidence learning~\cite{razlighi2025confidentsplatting}, and joint optimization frameworks~\cite{lee2026gaussianpopprincipledsimplificationframework}, we propose a strictly camera-agnostic pruning strategy that operates entirely on the learned Gaussian representation. 
	
	Conceptually, our method follows a two-stage formulation. First, importance is inferred from intrinsic splat-aware neighborhood descriptors computed directly from Gaussian parameters, without rendering supervision. Second, these descriptor signals are interpreted through a probabilistic Beta evidence model to estimate confidence and uncertainty for each splat. This separation between intrinsic structural measurement and statistical evidence interpretation enables one-shot pruning without retraining, recovery optimization, or camera visibility signals.

	\section{Proposed Method}
\begin{figure}[t]
		\begin{center}
			\fbox{ 
				\includegraphics[height=4.5cm, width=10cm]{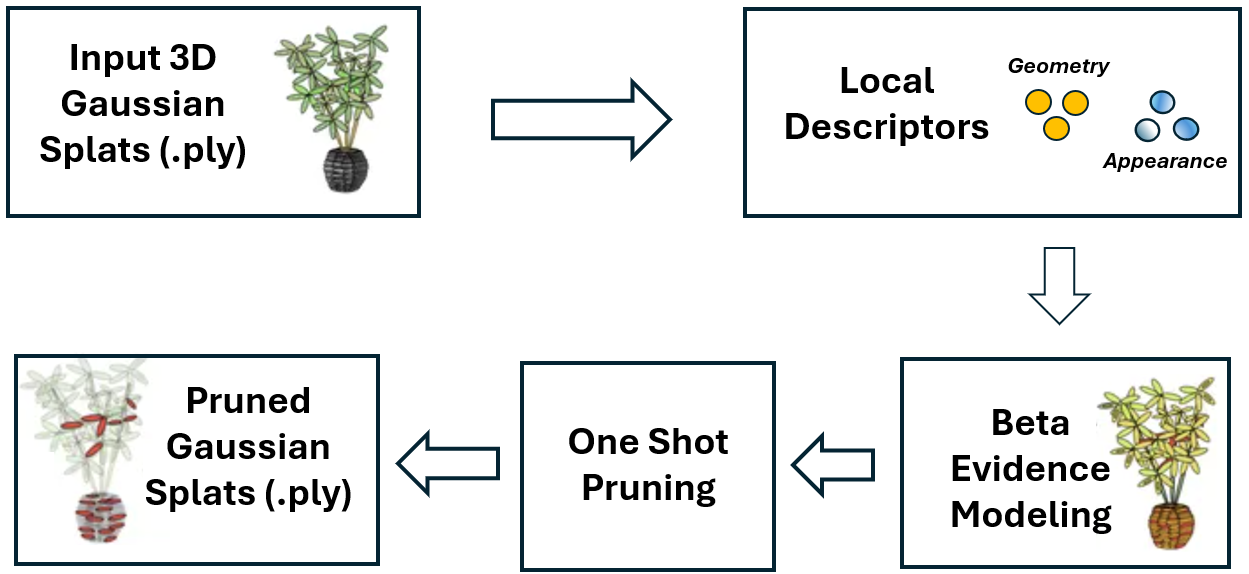}}
		\end{center}
		\caption{
			Method overview of the proposed camera-agnostic Gaussian splat pruning framework. 
			Given trained 3D Gaussian splats, local descriptor statistics are computed and transformed into Beta-distributed evidence to model confidence and uncertainty. 
			A one-shot confidence-aware ranking then removes redundant splats, balancing expected pruning likelihood and uncertainty to produce the final pruned representation.
		}
		\label{fig:method_overview}
	\end{figure}

	The overall pipeline of the proposed method is illustrated in Figure~\ref{fig:method_overview}. Given a trained set of Gaussians, our method proceeds in three stages: (1) computation of splat-aware neighborhood descriptors that capture intrinsic geometric and appearance consistency; (2) probabilistic confidence estimation via a Beta evidence model derived from descriptor statistics; and (3) one-shot ranking and pruning using an uncertainty-aware confidence criterion.
	
	%The entire process operates directly on the learned Gaussian parameters without access to cameras, rendered images, or training-time gradients.
	
	%The framework is lightweight, deterministic, and modular, and can be applied as a drop-in post-processing step to any 3D Gaussian splatting pipeline. 

	\subsection{Local Descriptor Extraction using HSFH}\label{sec:desc}
	
	For a trained 3DGS representation, individual splat parameters are often insufficient for robust reliability inference. Instead, reliability is assessed from local neighborhood 	consistency. For each splat $g_i$, we compute a Gaussian splat descriptor $\mathcal{D}_i$ over its spatial neighborhood $\mathcal{N}(i)$, capturing geometric and appearance relationships among
	nearby splats. Each splat is characterized by its position, covariance, opacity, and spherical harmonics appearance coefficients.
	
	To effectively capture splat reliability, we introduce the HSFH, a descriptor tailored to Gaussian splatting representations. HSFH extends the FPFH descriptor~\cite{rusu2009fpfh} by incorporating appearance cues derived from spherical harmonics.
	
	\paragraph{Geometric component.}
	The geometric component encodes local shape using histograms of Darboux
	frame angles $(\alpha, \sigma, \theta)$, as in FPFH. For each splat, the
	angles between surface normals and relative neighbor directions are
	quantized into three 11-bin histograms, which are concatenated to form a
	33-dimensional geometric descriptor. This representation captures local
	curvature, anisotropy, and relative orientation, and is robust to noise
	and irregular point distributions.
	
	\paragraph{Appearance component from spherical harmonics.} Appearance cues are derived from the spherical harmonics coefficients associated with each splat using two complementary encodings:
	
	1. \textit{Power spectrum $P_\ell$.}  
	Band-wise energy is computed for each spherical harmonics degree $\ell$, which provides a compact summary of radiance variation across angular frequencies.
	\[
	P_\ell = \sum_{m=-\ell}^{\ell} \| c_{\ell m} \|^2,
	\]

	2. \textit{Histogram representation.}  
	Spherical harmonics coefficients (or reconstructed color responses) are discretized into fixed bins to capture the distribution of local appearance variation within the neighborhood.
	
	\vspace{0.2cm}
	
	\noindent
	Both encodings are normalized and concatenated with the geometric descriptor to form the final camera-agnostic HSFH representation.
	
	\paragraph{Optional view-aware features.}
	When camera information is available, HSFH can be augmented with a low-dimensional component describing splat-view alignment. For each splat and camera, we compute: (i) the Euclidean distance between
	the splat center and the camera center, (ii) the cosine of the angle between the splat normal and the viewing direction, and (iii) the alignment between the splat orientation and the camera forward axis.
	These quantities are concatenated into a compact 10-dimensional vector that captures relative distance, angular alignment, and orientation.
	
	In strictly camera-agnostic settings, this view-aware component is omitted, and HSFH reduces to geometry and appearance features only. When available, view-aware cues provide additional retention evidence for grazing-angle splats and fine structures. %, but are not required for effective pruning.

	\subsection{Beta Evidence Modeling}\label{sec:beta}
	
	We model the splat pruning as a probabilistic decision under
	uncertainty. For each splat $g_i$, we define a Bernoulli random variable
	indicating whether the splat should be pruned, and model uncertainty using
	a Beta distribution,
	\begin{equation}
		p_i \sim \mathrm{Beta}(A_i, B_i),
		\label{eqn:simbeta}
	\end{equation}
	where $A_i$ represents evidence supporting \emph{retention} of splat $i$
	and $B_i$ represents evidence supporting \emph{safe pruning}. Beta
	distributions provide a principled representation of uncertainty for
	binary decisions and are widely used in uncertainty-aware learning and
	decision-making~\cite{kendall2017uncertainties}.
	
	The posterior mean
	\begin{equation}
		\mu_i = \frac{B_i}{A_i + B_i}
	\end{equation}
	corresponds to the expected pruning probability, while the variance
	\begin{equation}
		\sigma_i^2 = \frac{A_i B_i}{(A_i + B_i)^2 (A_i + B_i + 1)}
	\end{equation}
	captures the associated uncertainty.
	
	\paragraph{Descriptor-derived statistics and evidence mapping.}
	
	For each splat $g_i$, we compute a set of normalized scalar statistics
	$s_i,\ell_i,o_i,u_i \in [0,1]$ derived from descriptor aggregation over
	its spatial neighborhood $\mathcal{N}(i)$. These statistics summarize
	local geometric and appearance consistency:
	
	\begin{itemize}
		\item $\ell_i$ measures \emph{low geometric contrast}, computed as the
		inverse normalized standard deviation of the geometric descriptor
		block, such that homogeneous neighbourhoods yield high values.
		\item $s_i$ measures \emph{low-frequency appearance consistency},
		computed as the inverse normalized variance of appearance-related
		descriptor components.
		\item $o_i$ denotes the normalized splat opacity.
		\item $u_i$ denotes a \emph{geometry uniqueness} score, computed from
		the variability of descriptor components across dimensions, with
		higher values indicating structurally distinctive neighbourhoods.
	\end{itemize}

	These statistics are accumulated as soft evidence in the Beta model using distance-weighted aggregation in equation (\ref{eqn:evidencecollect}), where $w$ denotes an aggregation weight.
	\begin{equation}
		\begin{split}
			B_i &\leftarrow B_i + w\!\left(0.50\,s_i + 0.35\,\ell_i + 0.20(1-o_i)\right) + 0.20(1-u_i),\\
			A_i &\leftarrow A_i + w\!\left(0.55\,o_i\right) + 0.50\,u_i,
		\end{split}
		\label{eqn:evidencecollect}
	\end{equation}

	The weighting coefficients $(0.50, 0.35, 0.20)$ for $B_i$ and $(0.55, 0.50)$ for $A_i$ are chosen empirically and reflect the relative importance of geometric homogeneity, appearance consistency, opacity, and structural distinctiveness. As shown in Sec.~\ref{sec:sensitivity}, the method is not brittle to moderate perturbations of these coefficients: swapping preset weights around the default yields a PSNR spread of only $0.18$\,dB (and an SSIM spread of $0.002$) across four presets. Homogeneous neighbourhoods with low geometric and appearance variation increase pruning evidence, while opaque and structurally distinctive splats increase retention evidence. The formulation is flexible and can be tuned to emphasize geometry, appearance, or opacity depending on the target application.
	
	\paragraph{Optional camera-aware extensions.}
	
	When camera information is available, an additional grazing or edge-likeness proxy $e_i$ can be computed from normal-view alignment and opacity statistics. This term provides supplementary retention evidence
	for splats located near silhouettes, contours, or grazing-angle configurations. 
	
	In strictly camera-agnostic settings, this term is
	omitted. Importantly, effective pruning is achieved without any view-dependent information, and optional camera-aware cues serve only as auxiliary refinements when available.

	\subsection{One-Shot Pruning}\label{sec:OneShotPruning}
	
	A pruning decision can be derived from the lower confidence bound (LCB)
	of the Beta posterior in equation (\ref{eqn:betaconfi}) where $q$ is a user-defined quantile.
	
	\begin{equation}
		\mathrm{LCB}_i = \mathrm{BetaInvCDF}(q, A_i, B_i),
		\label{eqn:betaconfi}
	\end{equation}

	In practice, we approximate the LCB
	using a Gaussian moment form in equation (\ref{eqn:gaussian}), where $\mu_i$ and $\sigma_i$ denote the posterior mean and standard deviation,
	and $z$ controls the confidence level. 
	\begin{equation}
		\mathrm{LCB}_i \approx \mu_i - z\,\sigma_i.
		\label{eqn:gaussian}
	\end{equation}
	
	However, the formulation in equation (\ref{eqn:gaussian}) is pessimistic, as it penalizes uncertainty by lowering the score of high-variance components. When applied to pruning, such a rule
	removes uncertain coefficients early and therefore results in aggressive pruning. To avoid over-pruning fine structures and semi-transparent regions, we
	employ an \emph{optimistic confidence} inspired by the upper confidence bound (UCB) principle and shown in equation (\ref{eqn:prune}), where $\gamma$ controls the confidence level and is defined by the user. This \emph{score} softly rewards uncertainty and empirically stabilizes PSNR and SSIM under a high pruning ratio. 
	
	\begin{equation}
		\mathrm{score}_i = \mu_i + \gamma\,\sigma_i,
		\label{eqn:prune}
	\end{equation}

	\vspace{0.5cm}
	
	Finally, pruning is applied once by thresholding the confidence score using a condition in equation (\ref{finalthresh}), where $\tau$ is the threshold defined by the user and initialized in the next section. 
	\begin{equation}
		g_i \;\text{is pruned if}\; \mathrm{score}_i < \tau.
		\label{finalthresh}
	\end{equation}

	\section{Experiments}
	In this section, we evaluate the proposed camera-agnostic pruning method on multiple MPEG CTC test sequences. All experiments are conducted strictly \emph{post-training}, using the exported Gaussian splat representations (\texttt{.ply}). Our evaluation isolates the impact of two core components: (i) descriptor-based neighbourhood modelling and (ii) probabilistic Beta evidence accumulation. We perform systematic comparison against two camera-dependent state-of-the-art pruning baselines, an ablation study on each component, and dedicated sensitivity and runtime analyses (Sec.~\ref{sec:sensitivity}).

	\vspace{0.2cm}
	
	\paragraph{Evaluation protocol.}
	We follow the MPEG CTC evaluation protocol. All reconstruction metrics are computed against captured ground-truth images on held-out camera views that are not used during pruning or optimization. We report PSNR (dB), SSIM, and LPIPS (VGG backbone) at 1024\,px longest side across 14 CTC scenes, yielding 527 image pairs per condition. 
	
	\vspace{0.2cm}
	
	For controlled comparison, we construct pruning-only variants of two camera-dependent frameworks by isolating the pruning stages of LightGaussian~\cite{fan2024lightgaussiann} and Confident Splatting~\cite{razlighi2025confidentsplatting}, excluding additional compression components such as quantization or recovery optimization. These reduced variants are denoted \textit{LightGSPrune} and \textit{ConfSplatPrune}. Both baselines rely on image-based supervision and therefore operate with access to optimization viewpoints. To avoid potential evaluation bias, all methods are evaluated exclusively on held-out views following the MPEG CTC protocol.
	
	\subsection{Implementation Details}
	
	Pruning decisions are derived from descriptor-level statistics rather than raw Gaussian parameters. We compute local descriptors using HSFH (Section~\ref{sec:desc}) and estimate probabilistic confidence from aggregated descriptor statistics via the Beta evidence model (Section~\ref{sec:beta}). For efficiency on large-scale scenes, descriptor computation is accelerated with voxelized downsampling. Unless otherwise specified, voxel size is set to $1$--$2\%$ of the scene bounding-box diagonal, providing a practical trade-off between computational cost and local-structure fidelity. Descriptors and statistics are computed at voxel level and interpolated to individual splats by distance-based weighting, while final pruning decisions are applied at full splat resolution.
	
	Pruning is controlled by confidence threshold $\tau$ and uncertainty weight $\gamma$ (Section~\ref{sec:OneShotPruning}). We set $\tau$ by percentile thresholding to match a target pruning ratio (for example, $\tau \approx 0.30$ and $\tau \approx 0.10$ yield approximately $30\%$ and $10\%$ pruning). The uncertainty weight is fixed to $\gamma = 0.25$ in all experiments.
	
	\subsection{Quantitative Comparison}
	
	We evaluate the proposed camera-agnostic pruning method against two state-of-the-art baselines, \textit{LightGSPrune} and \textit{ConfSplatPrune}, derived from LightGaussian~\cite{fan2024lightgaussiann} and Confident Splatting~\cite{razlighi2025confidentsplatting}, respectively. All methods are evaluated on held-out MPEG CTC views under the standardized evaluation protocol.
	
	\vspace{0.2cm}
	
	Results in Table~\ref{table:comparison_pruning_levels} show that at low and medium pruning levels, camera-aware methods achieve the highest reconstruction fidelity due to view-dependent supervision. However, the proposed camera-agnostic approach remains competitive without access to image projections. At high pruning ratios, our method achieves slightly better performance on \textit{breakfast (tracked)} and \textit{cinema (tracked)} sequences, which exhibit complex spatial layouts and diverse viewpoint coverage. These large-scale scenes contain substantial spatial redundancy, making high-ratio pruning more meaningful and allowing descriptor-driven confidence modelling to effectively remove structurally redundant splats while preserving geometry. In contrast, the object-centric \textit{plant} sequence shows near-saturated performance across all methods, reflecting its compact geometry and limited splat count. Overall, descriptor-driven neighbourhood modelling coupled with uncertainty-aware confidence estimation enables effective structure-preserving pruning without view-dependent supervision, particularly in large and spatially diverse environments.

	\begin{table}[t]
		
		\begin{center}
			
			\resizebox{0.93\linewidth}{!}{%
				\begin{tabular}{|cc|ccc|ccc|ccc|}
					\hline
					Dataset & Metric
					& \multicolumn{6}{c|}{Camera-Aware Methods}
					& \multicolumn{3}{c|}{Our Method} \\
					\cline{3-11}
					
					&
					& \multicolumn{3}{c|}{ \textit{ConfSplatPrune} }
					& \multicolumn{3}{c|}{ \textit{LightGSPrune} }
					& \multicolumn{3}{c|}{ \textit{BetaDescPrune} } \\
					\cline{3-11}
					
					&
					& ~Low~ & ~Mid~ & ~High~
					& ~Low~ & ~Mid~ & ~High~
					& ~Low~ & ~Mid~ & ~High~ \\
					\hline
					
					\multirow{3}{*}{\shortstack{bartender \\ (tracked)}}
					& PSNR (↑) & 34.84 & 31.03 & 30.10 & 34.80 & 31.15 & 30.05 & 33.50 & 31.03 & 30.00 \\
					& SSIM (↑) & 0.991 & 0.990 & 0.985 & 0.992 & 0.990 & 0.983 & 0.990 & 0.982 & 0.973 \\
					& LPIPS (↓) & 0.015 & 0.028 & 0.031 & 0.015 & 0.029 & 0.030 & 0.019 & 0.034 & 0.049 \\
					\hline
					
					\multirow{3}{*}{\shortstack{breakfast \\ (tracked)}}
					& PSNR (↑) & 34.01 & 29.13 & {\bf 25.07} & 33.03 & 29.12 & {\bf 24.72} & 31.12 & 28.24 & {\bf 26.12} \\
					& SSIM (↑) & 0.992 & 0.981 & {\bf 0.946} & 0.991 & 0.981 & {\bf 0.945} & 0.988 & 0.975 & {\bf 0.962} \\
					& LPIPS (↓) & 0.021 & 0.026 & {\bf 0.057} & 0.020 & 0.027 & {\bf 0.060} & 0.015 & 0.029 & {\bf 0.043} \\
					\hline		
					
					\multirow{3}{*}{\shortstack{cinema \\ (tracked)}}
					& PSNR (↑) & 31.95 & 31.23 & {\bf 23.09} & 32.06 & 31.49 & {\bf 23.12} & 29.07 & 26.34 & {\bf 24.92} \\
					& SSIM (↑) & 0.989 & 0.987 & {\bf 0.940} & 0.990 & 0.989 & {\bf 0.942} & 0.982 & 0.969 & {\bf 0.954} \\
					& LPIPS (↓) & 0.011 & 0.016 & {\bf 0.066} & 0.010 & 0.014 & {\bf 0.066} & 0.022 & 0.040 & {\bf 0.057} \\
					\hline
					
					\multirow{3}{*}{plant}
					& PSNR (↑) & 42.05 & 42.05 & 42.01 & 42.06 & 42.05 & 41.99 & 41.72 & 38.26 & 35.58 \\
					& SSIM (↑) & 0.999 & 0.999 & 0.998 & 0.999 & 0.999 & 0.998 & 0.999 & 0.998 & 0.997 \\
					& LPIPS (↓) & 0.001 & 0.001 & 0.001 & 0.001 & 0.001 & 0.001 & 0.001 & 0.002 & 0.003 \\
					\hline					
				\end{tabular}
			} 
		\end{center}
		
		\caption{
			Comparison with prior pruning methods at low ($10\%$), medium ($20\%$), and high ($30\%$) pruning ratios on four MPEG CTC scenes. All metrics are computed between the pruned render and the non-pruned render (PSNR in dB, higher PSNR/SSIM better, lower LPIPS better). Methods are evaluated on held-out MPEG CTC camera views.
		}
		\label{table:comparison_pruning_levels}
	\end{table}

	\subsection{Ablation Study}
	
	To evaluate the contribution of individual components, we compare four pruning configurations: (i) pruning without descriptors or Beta modelling, (ii) pruning without descriptors but with Beta modelling, (iii) pruning with descriptors but without Beta modelling, and (iv) the full method combining descriptors with Beta evidence accumulation (see Table \ref{table:ablation_beta_descriptor}). Across all evaluated datasets, descriptor-based neighbourhood modelling improves pruning stability by incorporating local structural context beyond individual splat parameters. Incorporating Beta evidence modelling further enhances robustness through uncertainty-aware confidence estimation, reducing over-pruning in ambiguous regions. The full method consistently achieves the best reconstruction quality, demonstrating that structural descriptors and probabilistic uncertainty modeling provide complementary benefits for camera-agnostic pruning.

	\begin{table}[t]
		\centering
		
		\resizebox{0.6\linewidth}{!}{%
			\begin{tabular}{|cc|cc|cc|}
				\hline
				Dataset & Metric
				& \multicolumn{2}{c|}{No Descriptors}
				& \multicolumn{2}{c|}{With Descriptors} \\
				\cline{3-6}
				
				&
				& No Beta & Beta
				& No Beta & Beta \\
				\hline
				
				\multirow{3}{*}{\shortstack{bartender \\ (tracked)}}
				& PSNR (↑)   & 24.91 & 29.23 & 23.88 & \textbf{31.07} \\
				& SSIM (↑)   & 0.955 & 0.979 & 0.948 & \textbf{0.982} \\
				& LPIPS (↓)  & 0.055 & 0.035 & 0.074 & \textbf{0.034} \\
				\hline
				
				\multirow{3}{*}{\shortstack{breakfast \\ (tracked)}}
				& PSNR (↑)   & 19.98 & 28.90 & 17.94 & \textbf{34.47} \\
				& SSIM (↑)   & 0.933 & 0.978 & 0.903 & \textbf{0.979} \\
				& LPIPS (↓)  & 0.072 & 0.027 & 0.097 & \textbf{0.025} \\
				\hline
				
				\multirow{3}{*}{\shortstack{cinema \\ (tracked)}}
				& PSNR (↑)   & 21.76 & 27.15 & 19.45 & \textbf{27.16} \\
				& SSIM (↑)   & 0.937 & 0.970 & 0.925 & \textbf{0.972} \\
				& LPIPS (↓)  & 0.075 & 0.037 & 0.091 & \textbf{0.036} \\
				\hline
				
				\multirow{3}{*}{plant}
				& PSNR (↑)   & 34.73 & 38.05 & 34.73 & \textbf{38.05} \\
				& SSIM (↑)   & 0.995 & 0.998 & 0.995 & \textbf{0.998} \\
				& LPIPS (↓)  & 0.005 & 0.002 & 0.005 & \textbf{0.002} \\
				\hline
				
			\end{tabular}
			
		}
		\caption{Ablation over descriptor use and Beta modelling, at mid pruning ($20\%$). All metrics are computed between the pruned and non-pruned renders (higher PSNR/SSIM better, lower LPIPS better). Bold marks the best value per row.}
		\label{table:ablation_beta_descriptor}
	\end{table}

	\subsection{Robustness, Sensitivity, and Practical Deployment}
	\label{sec:sensitivity}
	
	To validate that the proposed method generalizes across diverse pruning intensities and remains robust to hyperparameter variations, we conducted a comprehensive set of robustness studies on the MPEG CTC dataset. These analyses address three key deployment concerns: (i)~fidelity at higher pruning ratios beyond the standard $30\%$ regime, (ii)~sensitivity to design choices such as aggregation coefficients and voxel sizes, and (iii)~computational overhead relative to the baseline. We also present per-scene analysis to characterize when descriptor-driven pruning excels and where it shows limitations.
	
	\subsubsection*{Reconstruction quality at higher pruning ratios}
	
	We evaluate performance across a comprehensive range of pruning intensities on the $14$-scene evaluation subset, following the MPEG CTC protocol with all held-out camera views at $1024$\,px longest side, yielding $527$ image pairs per condition. To characterize performance beyond the standard $30\%$ deployment regime, we evaluate descriptor contributions (with vs.\ without descriptors) across three keep-fractions: $0.7$ (conservative $30\%$ pruning), $0.5$ ($50\%$ pruning), and $0.3$ (aggressive $70\%$ pruning), as shown in Table~\ref{table:absolute_keep_fractions}.
	
	Quality degradation across pruning ratios is monotonic and gradual. PSNR declines smoothly with heavier pruning, SSIM remains above $0.87$ even at $70\%$ removal, and perceptual loss (LPIPS) shows no sudden artifacts. Notably, the descriptor performance gap remains modest even at extreme pruning ($\lesssim 1.6$\,dB PSNR, $\lesssim 0.012$ LPIPS), confirming that descriptor-driven pruning selects \emph{which} splats to remove more intelligently rather than introducing a quality cliff or degradation pattern.
	
	\begin{table}[t]
		\centering
		\resizebox{0.92\linewidth}{!}{%
			\begin{tabular}{|c|ccc|ccc|c|}
				\hline
				Prune \% & \multicolumn{3}{c|}{No Descriptors} & \multicolumn{3}{c|}{With Descriptors} & $\Delta$PSNR \\
				\cline{2-4}\cline{5-7}
				& PSNR ($\uparrow$) & SSIM ($\uparrow$) & LPIPS ($\downarrow$) & PSNR ($\uparrow$) & SSIM ($\uparrow$) & LPIPS ($\downarrow$) & (dB) \\
				\hline
				$30\%$  & 35.75 & 0.967 & 0.039  & \textbf{37.38} & \textbf{0.974} & \textbf{0.030} & $-1.63$\\
				$50\%$ & 32.79 & 0.944 & 0.061   & \textbf{33.68} & \textbf{0.948} & \textbf{0.054} & $-0.89$\\
				$70\%$  & 28.05 & 0.871 & 0.115  & \textbf{29.07} & \textbf{0.882} & \textbf{0.103} & $-1.02$\\
				\hline
		\end{tabular}}
		\caption{Pruning-fidelity across three keep-fractions on the $14$-scene MPEG CTC evaluation subset ($527$ image pairs per cell; all held-out views at $1024$\,px longest side). Same evaluation criteria as Tables~\ref{table:comparison_pruning_levels}--\ref{table:ablation_beta_descriptor}: PSNR in dB, SSIM in $[0,1]$, LPIPS (VGG). }
		\label{table:absolute_keep_fractions}
	\end{table}

	\subsubsection*{Per-scene variability and when descriptors help most}
	
	While the average performance difference between with and without descriptors remains modest (within $1.6$\,dB PSNR across all tests), per-scene analysis of the $14$-scene evaluation set reveals heterogeneous behavior. On several challenging scenes, \textit{solo\_tango\_male} produces (largest single-scene gain of $+4.04$\,dB PSNR at $70\%$ pruning), \textit{tango\_duo}, \textit{cinema (tracked)}, and \textit{breakdance (untracked)} descriptor guidance delivers substantial improvements in specific pruning regimes. Conversely, on repetitive fine-detail scenes such as \textit{flowerdance} (largest loss of $-12.4$\,dB at $70\%$ pruning), descriptor evidence can over-protect background splats, leading to sub-optimal pruning. This heterogeneous behavior is expected given descriptor-driven pruning operates without view-dependent supervision; descriptor effectiveness is fundamentally tied to spatial redundancy and appearance consistency. Practitioners should expect scene-dependent performance rather than uniform gains, though the method remains competitive on average.
	
	\subsubsection*{Robustness to design choices}
	
	A key concern for practical deployment is sensitivity to design hyperparameters. We conducted coefficient sensitivity analysis on four preset configurations: baseline ($B$-coefficients: $0.50, 0.35, 0.20$; $A$-coefficients: $0.55, 0.50$), flat, opacity-heavy, and structure-heavy. Evaluation was performed on a $5$-scene subset at keep $=0.7$ (conservative pruning). PSNR result ranges across presets span $36.26$ to $36.44$\,dB (spread of $0.18$\,dB) with SSIM between $0.9721$ and $0.9737$ (spread of $0.002$), with the default baseline sitting mid-range. This narrow spread demonstrates robustness to moderate perturbations of the evidence accumulation weights.
	
	Similarly, voxel size for descriptor aggregation represents a soft hyperparameter with practical implications for computational cost. On the same $5$-scene subset (keep $=0.7$, with descriptors), evaluating voxel sizes of $0.10$, $0.20$, and $0.30$ times the bounding-box diagonal yields flat PSNR results of $36.24$, $36.34$, $36.14$\,dB and SSIM of $0.9719$, $0.9724$, $0.9719$ respectively. This insensitivity to voxelization offers practitioners flexibility without affecting pruning quality.

	\subsubsection*{Computational cost}
	
	To ensure the method is practical for deployment, we measured end-to-end runtime from the $14$-scene evaluation (single CPU thread, method timing only). Baseline pruning without descriptors averages $6.78$\,s per scene (range $2.93$--$16.42$\,s), while descriptor-inclusive pruning averages $7.29$\,s per scene (range $3.08$--$18.09$\,s). The descriptor aggregation adds approximately $0.5$\,s per scene, representing only $\sim 8\%$ computational overhead. This modest cost is justified given the improved pruning decisions and robustness gains demonstrated above, making the method practical for offline processing pipelines.
	
	\subsubsection*{Generalization to the full MPEG CTC suite}
	
	To validate broader generalization, we evaluated on the complete MPEG CTC set of $19$ sequences ($7$ forward-facing dynamic, $12$ object-centric). Forward-facing scenes achieve PSNR of $35.15$, $32.21$, $30.13$\,dB at $10\%$--$30\%$ pruning with SSIM $0.989$--$0.969$ and LPIPS $0.019$--$0.050$, showing stable cross-view behavior. Object-centric sequences reach PSNR of $42.95$, $39.52$, $35.30$\,dB with SSIM $0.999$--$0.997$ and LPIPS $0.001$--$0.003$, confirming descriptor benefits for compact scenes. The performance remains consistent across diverse scene types and pruning intensities.

	\subsection{Visual Analysis} 
\begin{figure}[t]
		\begin{center}
			\fbox{ 
				\includegraphics[height=7.5cm, width=12.25cm]{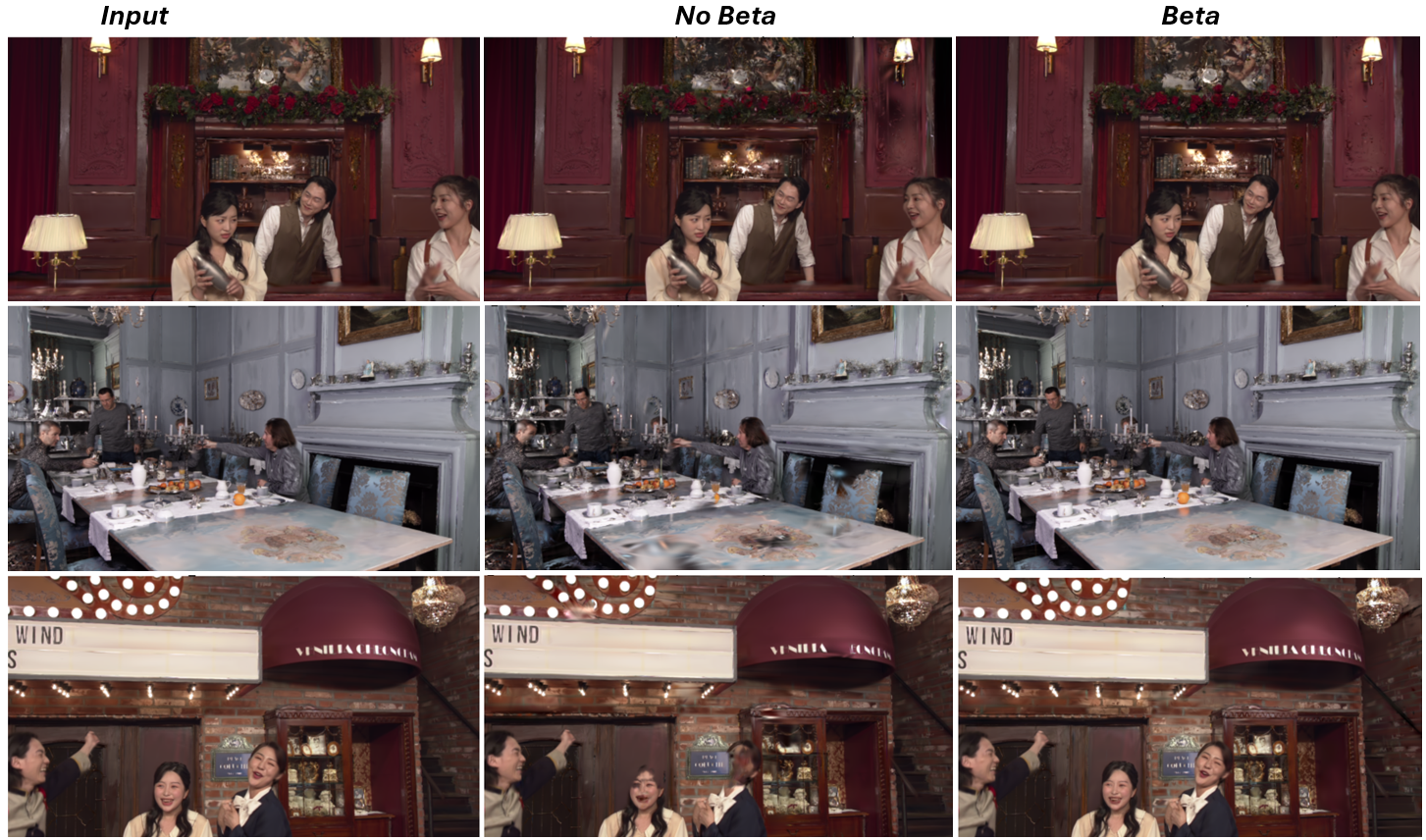}}
		\end{center}
		\caption{
			Qualitative comparison of camera-agnostic pruning on three MPEG CTC scenes (\textit{bartender (tracked)}, \textit{breakfast (tracked)}, \textit{cinema (tracked)}) at approximately 20\% pruning. The columns show (left to right) the input rendering, descriptor-only pruning without Beta modelling (No Beta), and uncertainty-aware pruning with Beta evidence (Beta). Incorporating Beta evidence improves preservation of fine structures and reduces visible pruning artefacts.
		}
		\label{fig:viewless_d}
	\end{figure}

	Figure~\ref{fig:viewless_d} shows qualitative results on three MPEG CTC scenes: \textit{bartender (tracked)}, \textit{breakfast (tracked)}, and \textit{cinema (tracked)}, at approximately 20\% pruning. The columns show the input rendering, descriptor-only pruning without Beta modelling (No Beta), and uncertainty-aware pruning with Beta evidence (Beta). The descriptor-only pruning removes redundant splats but introduces visible artefacts, particularly in thin structures, semi-transparent regions, and high-frequency geometric details. These appear as structural discontinuities and loss of fine geometry. Incorporating Beta evidence produces cleaner reconstructions with improved preservation of fine structures and more stable semi-transparent regions. Thus, accounting for uncertainty in the pruning decision leads to more consistent structure retention despite operating without camera information.

	These qualitative results align with the quantitative findings, showing that uncertainty-aware pruning improves visual fidelity and reduces pruning artefacts, particularly in geometrically complex regions.

	\section{Conclusion and Future Work}
	
	We presented a camera-agnostic, one-shot, post-training pruning framework for 3DGS based on descriptor-driven Beta evidence. By operating directly on exported Gaussian representations and eliminating reliance on camera parameters, rendered images, or training-time supervision, the proposed method enables robust and reproducible pruning in interchange-oriented settings such as the MPEG I-3DGS paradigm. The experiments on standard MPEG test sequences show that substantial redundancy can be removed while preserving reconstruction quality, even without view-dependent information. These findings demonstrate that reliable splat confidence can be inferred from intrinsic neighborhood structure alone and that effective post-training pruning does not require camera supervision. The one-shot and deterministic formulation further ensures low computational overhead, making the method well-suited for large-scale and standardized processing pipelines.
	
	\vspace{0.2cm}
	
	An important direction for future work is to investigate how descriptor-driven confidence estimation interacts with downstream compression and transmission stages. In particular, integrating confidence-aware pruning with geometry and attribute coding, adaptive quantization, and rate-distortion optimization offers a promising path toward improved end-to-end efficiency in immersive Gaussian splatting systems.

	%
	%\begin{figure}
	%\begin{tabular}{ccc}
	%\bmvaHangBox{\fbox{\parbox{2.7cm}{~\\[2.8mm]
				%\rule{0pt}{1ex}\hspace{2.24mm}\includegraphics[width=2.33cm]{images/eg1_largeprint.png}\\[-0.1pt]}}}&
	%\bmvaHangBox{\fbox{\includegraphics[width=2.8cm]{images/eg1_largeprint.png}}}&
	%\bmvaHangBox{\fbox{\includegraphics[width=5.6cm]{images/eg1_2up.png}}}\\
	%(a)&(b)&(c)
	%\end{tabular}
	%\caption{It is often a good idea for the first figure to attempt to
		%encapsulate the article, complementing the abstract.  This figure illustrates
		%the various print and on-screen layouts for which this paper format has
		%been optimised: (a) traditional BMVC print format; (b) on-screen
		%single-column format, or large-print paper; (c) full-screen two column, or
		%2-up printing. }
	%\label{fig:teaser}
	%\end{figure}
	%
	%
	%
	%\begin{figure*}
	%\begin{center}
	%\fbox{\rule{0pt}{2in} \rule{.9\linewidth}{0pt}}
	%\end{center}
	%   \caption{Example of a short caption, which should be centered.}
	%\label{fig:short}
	%\end{figure*}
	%
	%\begin{table}
	%\begin{center}
	%\begin{tabular}{|l|c|}
	%\hline
	%Method & Frobnability \\
	%\hline\hline
	%Theirs & Frumpy \\
	%Yours & Frobbly \\
	%Ours & Makes one's heart Frob\\
	%\hline
	%\end{tabular}
	%\end{center}
	%\caption{Results.   Ours is better.}
	%\end{table}

	\bibliography{egbib}
\end{document}